\documentclass[a4paper, 10pt, conference]{ieeeconf}      
\usepackage{FG2019}

\RequirePackage{fix-cm}
\usepackage{amssymb}
\usepackage{graphicx, subfigure}
\usepackage{multirow}
\usepackage{fancyhdr}
\usepackage{amsmath}
\usepackage{mdwmath}
\usepackage{fancyhdr}
\usepackage{mathtools}
\usepackage{eurosym}
\usepackage[utf8]{inputenc}
\usepackage{pifont}
\usepackage{bm}
\usepackage{cite}
\usepackage{xcolor}
\usepackage{amsfonts}
 
\usepackage{amssymb}
\usepackage{textcomp}
\usepackage[ruled,vlined,linesnumbered]{algorithm2e}
\usepackage{adjustbox}
\usepackage{url}

\FGfinalcopy 

\usepackage{color}

\IEEEoverridecommandlockouts                              
\def\FGPaperID{109} 

\title{\LARGE \bf
On the effect of age perception biases for real age regression
}

\begin{document}


\ifFGfinal
\thispagestyle{empty}
\pagestyle{empty}

\author{\parbox{16cm}{\centering
    {\large Julio C. S. Jacques Junior$^{1,2}$, Cagri Ozcinar$^3$, Marina Marjanovic$^4$, Xavier Baró$^{1,2}$, Gholamreza Anbarjafari$^{5,6}$ and Sergio Escalera$^{2,7}$}\\
    {\normalsize
    $^1$Universitat Oberta de Catalunya, Spain,
    $^2$Computer Vision Center, Spain,
    $^3$School of Computer Science and Statistics, Trinity College Dublin, Ireland,
    $^4$Singidunum University, Serbia,
    $^5$University of Tartu, Estonia,
    $^6$Institute of Digital Technologies, Loughborough University London, UK,
    $^7$University of Barcelona, Spain
    }}
    %
}


\else
\author{Anonymous FG 2019 submission\\ Paper ID \FGPaperID \\}
\pagestyle{plain}
\fi
\maketitle

\begin{abstract}
Automatic age estimation from facial images represents an important task in computer vision. This paper analyses the effect of gender, age, ethnic, makeup and expression attributes of faces as sources of bias to improve deep apparent age prediction. Following recent works where it is shown that apparent age labels benefit real age estimation, rather than direct real to real age regression, our main contribution is the integration, in an end-to-end architecture, of face attributes for apparent age prediction with an additional loss for real age regression. Experimental results on the APPA-REAL dataset indicate the proposed network successfully take advantage of the adopted attributes to improve both apparent and real age estimation. Our model outperformed a state-of-the-art architecture proposed to separately address apparent and real age regression. Finally, we present preliminary results and discussion of a proof of concept application using the proposed model to regress the apparent age of an individual based on the gender of an external observer.
\end{abstract}

\section{INTRODUCTION}

\begin{figure}[tbp]
\centering
\subfigure{\includegraphics[width=1.2cm]{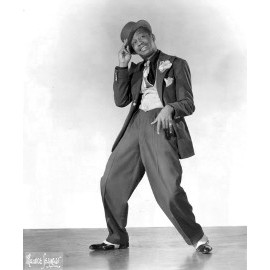}}
\subfigure{\includegraphics[width=1.2cm]{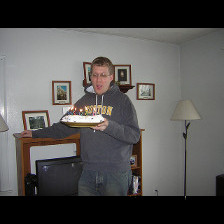}}
\subfigure{\includegraphics[width=1.2cm]{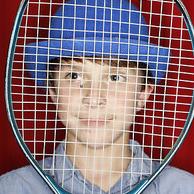}}
\subfigure{\includegraphics[width=1.2cm]{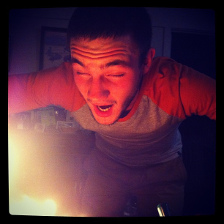}}
\subfigure{\includegraphics[width=1.2cm]{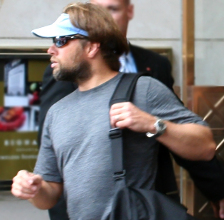}}
\subfigure{\includegraphics[width=1.2cm]{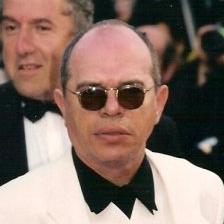}}
\subfigure{\includegraphics[width=1.2cm]{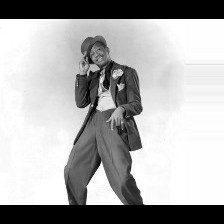}}
\subfigure{\includegraphics[width=1.2cm]{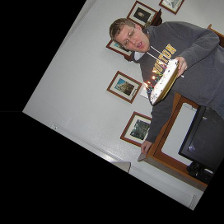}}
\subfigure{\includegraphics[width=1.2cm]{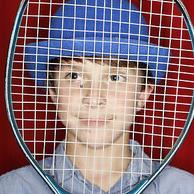}}
\subfigure{\includegraphics[width=1.2cm]{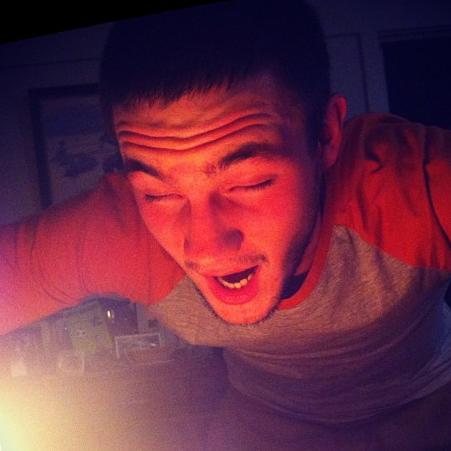}}
\subfigure{\includegraphics[width=1.2cm]{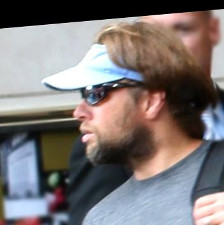}}
\subfigure{\includegraphics[width=1.2cm]{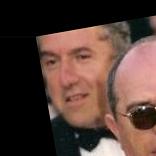}}
\subfigure{\includegraphics[width=1.2cm]{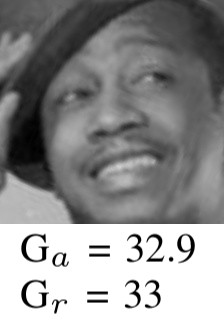}}
\subfigure{\includegraphics[width=1.2cm]{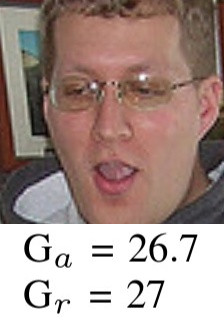}}
\subfigure{\includegraphics[width=1.2cm]{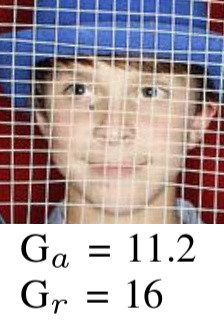}}
\subfigure{\includegraphics[width=1.2cm]{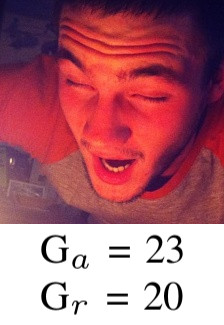}}
\subfigure{\includegraphics[width=1.2cm]{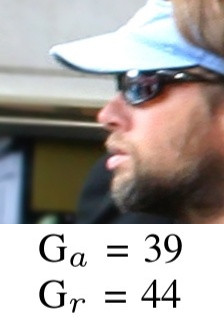}}
\subfigure{\includegraphics[width=1.2cm]{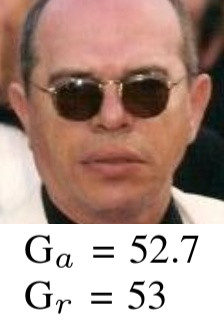}}
\caption{Samples of the APPA-REAL dataset~\cite{Agustsson2017} with respective apparent (G$_a$) and real (G$_r$) ground truth. First row: original images. Second row: respective ``cropped faces'' (provided with the dataset, using~\cite{Mathias:2014}). Third row: cropped faces obtained with~\cite{75Zhang:SPL:2016}, used in this work.}
\label{fig:dataset}
\end{figure}

Automatic age estimation from still images is intensively studied in Computer Vision~\cite{Angulu2018,clapes2018apparent,Agustsson2017} due to its wide range of possible applications, including forensics~\cite{Albert2007}, monitoring and surveillance~\cite{Dhimar:2016} (\textit{e.g.}, to search for a suspect with a specific age in a database), and recommendation systems~\cite{Alashkar:AAAI:2017} (\textit{e.g.}, ``\textit{how do I perceive to others?}''), just to mention a few. Age estimation task requires dealing with several factors such as human variations in appearance or head pose, the use of accessories (\textit{e.g.}, glasses, makeup), hair-style, as well as different illumination conditions, noise and/or occlusion~\cite{guo2009study}. Moreover, ageing is a variable-paced process depending on each person's genetics and other physiological factors, which make the task even more challenging~\cite{clapes2018apparent}. For the sake of illustration, few sample images used in this work containing both real and apparent age labels are shown in Fig.~\ref{fig:dataset}.

Recent research activities in machine learning (and deep learning) has started to focus on different aspects impacting the outcomes of automatic systems by taking into account subjectivity, human bias perception, fairness, and inclusiveness. In the case of age estimation, recent works started to analyse the apparent age~\cite{Agustsson2017} and the perceptual bias involved in age perception~\cite{clapes2018apparent}. With respect to person perception, recent studies proposed to analyse the biases introduced by observers opinion which are conditioned to facial attributes appearing in a given face image~\cite{Ryu2018,Mohsan2018}. More interestingly, even with the involved biases, the apparent age labels of face images are proved~\cite{Agustsson2017,clapes2018apparent} to achieve better performance for real age regression than training using real age labels.

The present work is inspired by the work of Clap\'es \textit{et al.}~\cite{clapes2018apparent}. They showed that using apparent age labels instead of real ones improves overall real age estimation. Furthermore, they presented preliminary analysis and discussion about the influence of subjective bias in age estimation, \textit{i.e.}, if we can estimate how much an attribute in the data influence/deviates from the target age, then we can correct final estimation and further improve the results. For instance, they post-processed obtained results and reduced the estimated real ages in a predefined manner after observing people (in general) overestimate female's age.

In this work, we present an end-to-end architecture for real and apparent age estimation which takes into account the way people in the images are perceived by external observers in a multi-task style scenario. We take benefit of apparent age estimation to improve the real age estimation by considering gender, facial expression, happiness and makeup levels\footnote{Attribute categories used in this work are imperfect for many reasons. For example, there is no gold-standard for ``race'' categories, and it is unclear how many race and gender categories should be stipulated (or whether they should be treated as discrete categories at all). We base on an ethical and legal setting to perform this work, and our methodology and bias findings are expected to be applied later to any re-defined and/or extended attribute category.}. While the end-to-end network in its first layers uses these face attributes as bias to improve apparent age estimation, the last layers of the network are in charge of doing the opposite, \textit{i.e.}, benefiting from an improved apparent age estimation and face attributes to unbias apparent predictions to regress the real age. We show that in practice this works better than directly regressing real age from real age labels (even if face attributes are considered).

Rather than focusing on outperforming the state-of-the-art on apparent/real age estimation~\cite{Agustsson2017}, our main goal is to validate the hypothesis that improvements in both apparent and real age estimation can be tackled jointly in an end-to-end fashion when combined with specific attributes people usually use in everyday life when drawing first impressions about others (\textit{e.g.}, gender, ethnic, facial expression). As far as we know, the work of Clap\'es \textit{et al.}~\cite{clapes2018apparent} is the only study proposed to deal with several attributes and apparent age to improve real age estimation. 

The contributions of our work are: (i) we provide with an automatic end-to-end approach to improve apparent age prediction and use this together with attributes to also improve real age estimation. Implicitly, it performs the following: in the first part of the network, attributes (bias) are used to improve apparent age estimation and in the second part, from apparent age to real age regression those same attributes are used but in an inverse way, to unbias the apparent age to approximate them to the real age value; (ii) we analyse individually the influence of adopted attributes for real and apparent age estimation; (iii) we outperformed the baseline model~\cite{clapes2018apparent} by using a more robust architecture trained end-to-end, rather than using an add-hoc post-processing stage; and (iv) we present preliminary results and discussion when the gender attribute of people who label the data is also taken into account. This way the model is able to regress the apparent age of an individual based on the gender of an external observer.


The rest of the paper is organised as follows: Section~\ref{sec:related} presents the related work on real and apparent age estimation, with a particular focus on face attributes analysis as a source of bias for age estimation. The proposed model and experimental results are presented in Sections~\ref{sec:model} and~\ref{sec:results}, respectively. Finally, our conclusions are drawn in Section~\ref{sec:conclusions}.

\section{Related work}
\label{sec:related}
This section reviews related work on real and apparent age estimation. Early and recent works are briefly discussed without the intention of providing an extended and comprehensive review on the topic. To this end, we refer the reader to~\cite{Angulu2018}. Then, we revisit related studies on the analysis of bias in age estimation.

\subsection{Real Age Estimation}
Early works on real age estimation are based on handcrafted approaches. Lanitis \textit{et al.}~\cite{Lanitis2004} proposed to use active appearance models to define compact feature spaces which are regressed for real age estimation. Yun \textit{et al.}~\cite{Fu2007} based on manifold analysis and multiple linear regression functions for real age estimation. Guodong~\textit{et al.}~\cite{guo2009study} analysed age estimation using a pyramid of Gabor filters.

Recent works for real age estimation in images and videos benefit from the advances in deep learning and end-to-end architectures. For instance, Pei~\textit{et al.}~\cite{pei2017attended} proposed an end-to-end architecture for learning the real age of given facial video sequence. 
The model is based on the extraction of latent appearance representations which are learnt by a Recurrent Neural Network (RNN). 
Gonz{\'a}lez-Briones \textit{et al.}~\cite{gonzalez2018multi} proposed an ensemble of age and gender recognition techniques, showing the benefits of late fusion from independent learners. 

\subsection{Apparent age estimation} 

In the case of apparent age estimation, each face image usually contains multiple age labels, related to variations in perception coming from different annotators/observers. 
Agustsson~\textit{et al.}~\cite{Agustsson2017} reported that  real age estimation could be successfully tackled as a combination of apparent and real age estimation by learning residuals. Geng \textit{et al.}~\cite{Geng2007} modeled an aging pattern by constructing a subspace given a set of ordered face images by age. In the aging pattern, each position indicates its apparent age. Zhu \textit{et al.}~\cite{Zhu2015} proposed to learn deep representations in a cascaded way. They analysed how to utilise a large number of face images without apparent age labels to learn a face representation, as well as how to tune a deep network using a limited number of labelled samples. 
Malli~\textit{et al.}~\cite{Malli2016} proposed to group face images within a specified age range to train an ensemble of deep learning models. The outputs of these trained models are then combined to obtain a final apparent age estimation.

\subsection{Analysis of bias in age estimation} 

While state-of-the-art machine learning algorithms can provide accurate prediction performances for age estimation, either if real or apparent age are considered, they are still affected by different variations in face characteristics. But how can age prediction performances be enhanced in this case? With this objective in mind, the analysis of bias in age perception has recently emerged~\cite{clapes2018apparent,Mohsan2018}. Can we better understand age perception and their biases so that the findings can be used to regress a better real age estimation? In this line, Clap{\'e}s~\textit{et al.}~\cite{clapes2018apparent} found some consistent biases in the APPA-REAL~\cite{Agustsson2017} dataset when relating apparent to real age. However, an end-to-end approach for bias removal was not considered. According to Alvi~\textit{et al.}~\cite{Mohsan2018}, training an age predictor on a dataset that is not balanced for gender can lead to gender biased predictions. They presented an algorithm to remove biases from the feature representation, as well as to ensure that the network is blind to a known bias in the dataset. Thus, improving classification accuracy, particularly when training networks on extremely biased datasets.

Differently from previous works, our aim is to explicitly use face attributes as sources of biases to regress a more accurate apparent age. Then, to unbias perceived age for real age regression. As we show later, taking benefit of biases in age perception can improve accuracy of real age estimation.

\section{Proposed Model}
\label{sec:model}

This section describes the proposed end-to-end architecture to jointly predict apparent and real age from facial images. The proposed architecture consists of two main stages. At the first stage, our goal is to better approximate the human perception mechanism by introducing human perception bias when predicting apparent ages using face attributes. 
At the second stage, we aim to remove bias when predicting the real age. The proposed model combines apparent and real age labels with additional face attributes (\textit{i.e.}, gender, race, level of happiness, and makeup) during training. 
Note that, once the model is trained, it uses neither real nor apparent age labels on the test set.

To achieve our objectives, and deal with the problem of jointly estimating apparent and real ages, we modified the VGG16 model~\cite{simonyan2014very}, which was pre-trained on ImageNet. The VGG16 model is a high capacity network utilised in most face analysis benchmarks. 
We modify the last layers of this base model in a way to reduce the bias of the apparent age estimation and accurately estimate the apparent and real age. Figs.~\ref{overview:baseline} and ~\ref{overview:proposed} illustrate the VGG16 base model with our modifications, which are explained next.


\subsection{Apparent age estimation}
\label{modelA}


To introduce human bias into the model, different attributes people use to perceive others are considered, which may affect the perceived age. They are gender (male, female), race (Asian, Afroamerican and Caucasian), level of happiness (happy, slightly happy, neutral and other) and makeup (makeup, no makeup, not clear and very subtle makeup). Note that state-of-the-art methods could be used to accurately recognise such attributes from face images (e.g.,~\cite{gonzalez2018multi, Corneanu_2018_ECCV, Wang:2018,Alashkar:AAAI:2017}). However, as the focus of our work is not on improving the recognition accuracy of such attributes, and because of the required amount of data to learn those associated tasks accurately, we decided to import them directly from the adopted dataset~\cite{clapes2018apparent}.

For the sake of simplicity, first consider the convolutional layer highlighted by a blue box in Fig.~\ref{overview:baseline} and Fig.~\ref{overview:proposed} (\textit{block5\_pool}) represent the same layer in both models, the base VGG16 and the proposed one. This illustrates which layers have been removed and which ones have been introduced. Such convolutional layer has high dimensionality, in particular if we consider the idea of combining with it a very low dimensional vector composed of people's attributes. In this work, each different attribute is encoded using one hot vector, usually employed to represent categorical data (\textit{e.g.}, male = [0, 1] and female~=~[1, 0]). All considered attributes are then concatenated, resulting in a vector (\textit{input\_2}, in Fig.~\ref{overview:proposed}) of length $13$, which is further encoded in a dense layer (\textit{hidden\_layer}, $10$D). In order to reduce the dimensionality of the previously mentioned high dimensional layer, a new ($7\times7\times512$) convolutional layer, followed by ReLU, is included just after it. The resulting (\textit{flatten\_1}) layer is then concatenated with people's attributes. Finally, a new FC layer ($fc2=256$D) is responsible to fuse both information before regress the apparent age, using a \textit{Sigmoid} function. 

With these network updates, we expect it to benefit from people's attributes to better approximate apparent age rather than just considering raw input image.

\subsection{Real age estimation}\label{modelR}

In the previous section we described how we perform apparent age estimation from visual information and people's attributes. As mentioned before, recent works~\cite{Agustsson2017,clapes2018apparent} reported that real age estimation could be better approximated when apparent age labels are used rather than real age ones. The main hypothesis behind such idea is that it could somehow help the network by reducing the ``noise" in the label space, and in particular of those people with a real age that highly differ to their apparent one.

In this work, instead of applying a post-processing bias correction scheme based on statistical analysis as in~\cite{clapes2018apparent}, we propose to correct such human perception bias during training. To achieve this goal, the predicted apparent age (represented by a ``salmon'' box in Fig.~\ref{overview:proposed}) is first concatenated with face perceptual attributes using a different representation (\textit{hidden\_layer\_2}, $5$D). Then, the concatenated information is fused in a FC layer ($fc3=6$D) before regressing the real age. 
This way, we expect real age estimation can benefit from both, apparent age estimation and people attributes.

\subsection{Training strategy}\label{sec:training}

We perform the training in two stages. In the first stage, just the included layers are fine-tuned, \textit{i.e.}, those shown in Fig.~\ref{overview:proposed}). In a second stage, the whole network is trained in an end-to-end manner. Adam algorithm is used as an optimisation method with default values, except the learning rate ($lr$), which was set based on the results obtained on validation. The mean squared error is defined as a loss function for both apparent and real age estimation, with both losses having equal weight.

Each case study presented in Sec.~\ref{sec:results} had its associated model trained using a train/validation sets and evaluated in a complete and different test set, all provided with the adopted APPA-REAL~\cite{Agustsson2017} database. We perform early stopping based on the validation set loss. The maximum number of epochs was set to $3000$ and $1500$, for stage 1 and 2, respectively. However, in most of the cases (except for Case 1, described next) the training stopped before achieving such limits. Anyway, no significant improvements have been observed on Case 1 with higher iterations. As stop criteria for predicting apparent age, the Mean Absolute Error (MAE) with respect to apparent age labels was adopted, whereas when predicting the real age the stop criteria considered the loss with respect to real age labels. Finally, the best model is kept based on the accuracy computed on the validation set.

   \begin{figure*}[htbp]
      \centering
      \includegraphics[height=2.3cm]{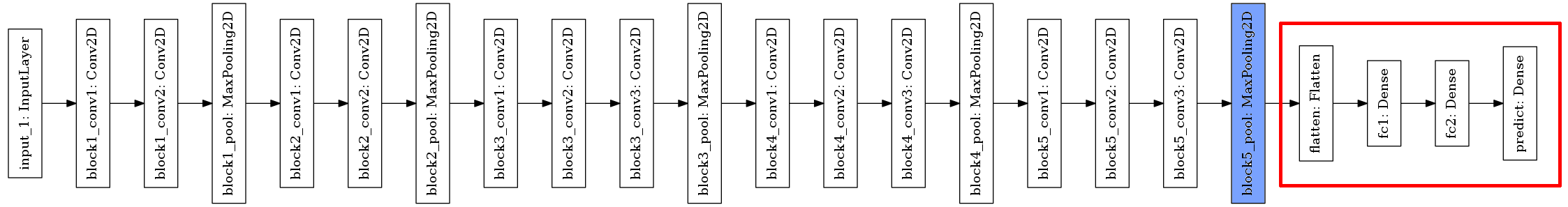}
      \caption{VGG16 base model. Layers highlighted in red have been modified to jointly predict apparent and real ages, as detailed in Fig.~\ref{overview:proposed}.}
      \label{overview:baseline}
   \end{figure*}
   
   \begin{figure}[htbp]
      \centering
      \includegraphics[height=5.5cm]{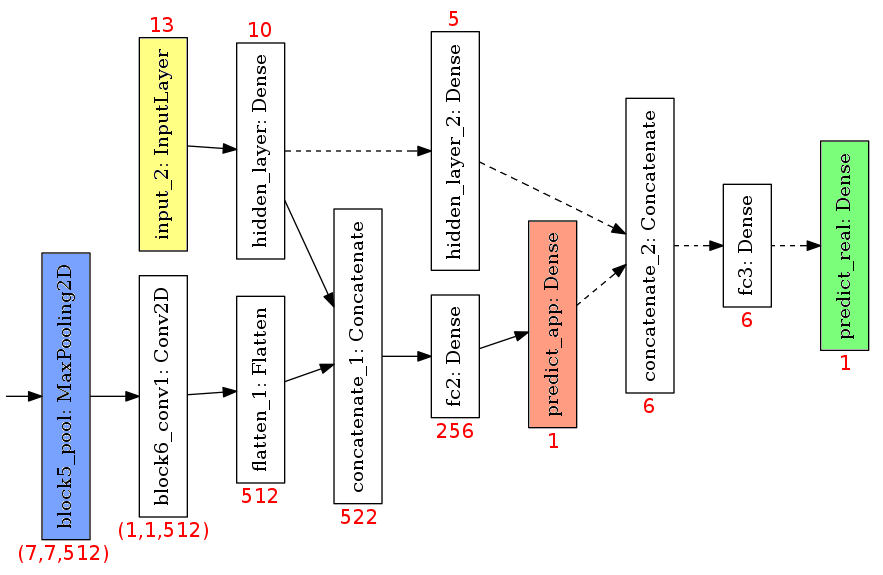}
      \caption{Overview of the proposed model. It uses the same structure as the  VGG16 base model from the \textit{input\_1} to the last convolutional layer (highlighted in blue, also shown in Fig.~\ref{overview:baseline}). Then, new layers are include to deal with people's attributes such as gender, race, happiness and makeup level (\textit{input\_2}, represented by a yellow box) before regressing the apparent age (``salmon'' box). The predicted apparent age is then used to support the final real age estimation (green box). The dashed arrows illustrates the connections responsible to fuse both apparent and real age information. The dimension of new layers are shown below/above them.}
      \label{overview:proposed}
   \end{figure}

\section{Experimental Results}\label{sec:results}

This section describes experimental results obtained in three case studies. The first case study (Sec.~\ref{sec:case1}) is based on the original VGG16 model, with the final layer modified to regress either apparent or real age. The second case study (Sec.~\ref{sec:case2}) illustrates the benefit of including people's attributes to the visual stream using a fraction of the proposed model, \textit{i.e.}, ignoring those dashed connections in Fig.~\ref{overview:proposed} and predicting either the apparent or the real age, based on the given input. The third case study (Sec.~\ref{sec:case3}) shows experimental results obtained using the proposed end-to-end architecture to jointly estimate apparent and real age. This way, we show improvements obtained with the inclusion of new features and more complex models incrementally. Then, a discussion on the results based on the considered people attributes is presented in Sec.~\ref{sec:attributes}. Finally, we show a proof of concept application using the proposed model to regress apparent age based on the gender of an external observer (Sec.~\ref{sec:workers}). Next, we briefly describe the adopted dataset and evaluation protocol.

\subsection{Dataset and evaluation protocol} 

The APPA-REAL~\cite{Agustsson2017} database was extended in~\cite{clapes2018apparent} with the inclusion of few attributes about people appearing in the images, \textit{i.e.}, gender (male, female), race (Asian, Afro-American and Caucasian), happiness (happy, slightly happy, neutral and other) and makeup category (makeup, no makeup, not clear and very subtle makeup). They also provide for a subset of the dataset the gender of people who labelled the data (used in Sec.~\ref{sec:workers}). The dataset is composed by almost 8K images of (mostly) single persons in frontal faces. However, it also includes images of full body, upper body or an expanded region of the face, with different resolutions (black and white or coloured images), different image qualities and severe head pose variation and occlusion problems. In some occasions there are multiple faces (mainly on the background), making the task even more challenging. The dataset is also provided with cropped and aligned faces, captured using~\cite{Mathias:2014}. However, some false positive samples, as well as wrongly detected faces, are still present after such procedure. To minimise the face detection related problems, we adopt a more robust face detection/alignment algorithm~\cite{75Zhang:SPL:2016}. A qualitative comparison about both approaches is illustrated in Fig.~\ref{fig:dataset}.


To quantitatively evaluate the results, we followed the evaluation protocol defined in~\cite{Agustsson2017,clapes2018apparent}, \textit{i.e.}, computing the Mean Absolute Error (MAE) between the ground truth and predicted ages on the test set.

\subsection{Case 1: VGG16 baseline}\label{sec:case1}
In this case study, the VGG base model is used as it is shown in Fig.~\ref{overview:baseline}. However, the last layer is modified to regress one single value using a \textit{Sigmoid} function. The model is trained as described in Sec.~\ref{sec:training}, using $lr_1=1e-6$. Obtained results for different inputs/outputs are shown in Table~\ref{table:baseline1}, top. As it can be seen, real age estimation was better predicted from apparent age labels than from real ones, which is aligned with results obtained in~\cite{clapes2018apparent} (see Table~\ref{table:results}). It emphasises the point that human bias introduced by external observers can be used to improve real age estimation. Next, we show how the inclusion of people's attributes can further improve these results.

\subsection{Case 2: (extended) VGG16 with face attributes}\label{sec:case2}
In this experiment, just a fraction of the proposed model is used. Concretely, those layers connected by dashed lines in Fig.~\ref{overview:proposed} have not been considered. Then, instead of predicting apparent age on the \textit{predict\_app} layer, and refining the real age from its output, we simply predicted either apparent or real age, according to the desired goal and respective input, \textit{i.e.}, apparent or real age labels. The model is trained as described in Sec.~\ref{sec:training}, using $lr_1=1e-4$. Obtained results for different inputs/outputs are shown in Table~\ref{table:baseline1}, bottom. As it can be seen, the inclusion of people's attributes helped to improve all scenarios (compared to case~1), aligned with the assumption that the way people perceive others is strongly influenced by gender~\cite{Mattarozzi:PLOS:2015}, facial expression~\cite{Sutherland:2016}, among other attributes. Indeed, this way real age estimation was able to benefit from both human biases introduced by external observers and people's attributes. 

To verify if the improvements obtained by the inclusion of people's attributes (\textit{i.e.}, case 2) were not influenced by the model architecture, compared to ``case 1'', we considered an additional experiment where ``case 2'' did not take into account people's attributes, \textit{i.e.}, the \textit{input\_2} layer shown in Fig.~\ref{overview:proposed} has not been considered (as well as its outputs). Obtained results, referred in Table~\ref{table:baseline1} as \textit{case 2'}, show that the inclusion of people's attributes still slightly benefit both apparent and real age estimation.

\begin{table}[htbp]
\caption{Case studies 1 and 2' (without face attributes), and 2 (with face attributes) results.}
\vspace{-0.2cm}
\label{table:baseline1}
\begin{center}
\begin{tabular}{|c|c|c|r|} \hline
 \textbf{Case study} & \textbf{Input label} & \textbf{Predict} & \textbf{MAE}  \\ \hline
 \multirow{3}{*}{1} 
 & App & App   &  7.532 \\ \cline{2-4}
 & App & Real  &  9.199 \\ \cline{2-4}
 & Real & Real &  10.385 \\ \hline
 \multirow{3}{*}{2'} 
 & App  - att  & App   &  6.228 \\ \cline{2-4}
 & App  - att  & Real  &  7.517 \\ \cline{2-4}
 & Real - att  & Real  &  7.909 \\ \hline
\multirow{3}{*}{2} 
 & App  + att  & App   &  6.024 \\ \cline{2-4}
 & App  + att  & Real  &  7.483 \\ \cline{2-4}
 & Real + att  & Real  &  7.782 \\ \hline
\end{tabular}
\end{center}
\end{table}

\subsection{Case 3: proposed model}\label{sec:case3}
In this case study, we report the results obtained using the (complete) proposed model\footnote{Code available at: \url{www.github.com/juliojj/app-real-age}} shown in Fig 3. The model is trained as described in Sec.~\ref{sec:training}, using $lr_1=1e-4$. Obtained results are shown in Table~\ref{table:results}, as well as those reported in~\cite{clapes2018apparent}. Note that our main goal is not to outperform the state-of-the-art in apparent/real age estimation, but to predict apparent and real ages jointly considering people's attributes in an end-to-end fashion. Thus, benefiting from human bias during training. As it can be seen, the proposed model outperformed~\cite{clapes2018apparent} and further improved apparent and real age estimation compared to previous cases 1 and 2, indicating that tackling the problem jointly benefit both apparent and real age regression tasks.

\begin{table}[htbp]
\caption{State-of-the-art Comparison.}
\label{table:results}
\vspace{-0.2cm}
\begin{center}
\begin{tabular}{|c|c|c|r|} \hline
\textbf{Model} & \textbf{Input label} & \textbf{Predict} & \textbf{MAE}  \\ \hline
\multirow{2}{*}{\cite{clapes2018apparent} } & App + att & Real &  13.577 \\ \cline{2-4}
                          & Real + att & Real & 14.572  \\ \hline
\multirow{2}{*}{Proposed} & \multirow{2}{*}{App + Real + att} & App & 6.131  \\ \cline{3-4}
                          & & Real & 7.356  \\ \hline
\end{tabular}
\end{center}
\end{table}

Table~\ref{table:parameters} shows the number of trainable parameters for each model used in previous experiments. As it can be seen, the proposed model was able to achieve better results than VGG16 baseline (Case 1) using a significantly smaller number of trainable parameters.

\begin{table}[htbp]
\caption{Number of trainable parameters.}
\label{table:parameters}
\vspace{-0.2cm}
\begin{center}
\begin{tabular}{|c|r|} \hline
\textbf{Model} & \textbf{Parameters} \\ \hline
Case study 1 &  134,264,641 \\ \hline
Case study 2 &  27,694,541 \\ \hline
Proposed model &  27,694,645 \\ \hline
\end{tabular}
\end{center}
\end{table}

\begin{figure}[htbp]
      \centering
      \includegraphics[height=4.2cm]{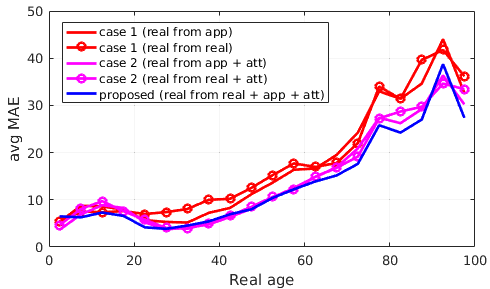}
      \caption{Real age estimation: average error distribution.}
      \label{error:real-curve}
   \end{figure}

\begin{figure}[htbp]
      \centering
      \includegraphics[height=4.2cm]{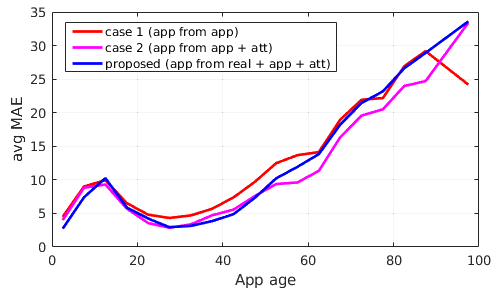}
      \caption{Apparent age estimation: average error distribution.}
      \label{error:app-curve}
   \end{figure}

\begin{figure}[htbp]
      \centering
      \includegraphics[height=2.8cm]{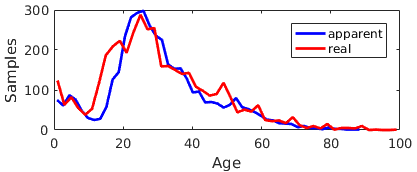}
      \caption{Apparent and real age distribution on the train set. }
      \label{hist:train-set}
   \end{figure}

Fig.~\ref{error:real-curve} shows the average error, with respect to real age estimation, for different age ranges obtained from different inputs and case studies (computed considering a window of five years to facilitate visualisation).  
As it can be seen, real age estimation overall benefited from people's attributes and apparent labels when different age ranges are considered. The proposed model (Case 3) obtained similar or better results for real age estimation compared to Case 2, indicating that the problem can benefit from both tasks (real and apparent age estimation) when jointly analysed. In Fig.~\ref{error:app-curve} a similar plot is shown but for apparent age estimation.
As it can be seen, the (complete) proposed model did not improve its partial version (Case 2) for some age ranges (\textit{e.g.}, higher than $\sim50$) with respect to apparent age estimation. This may be due to the fact that the proposed model was optimised (during training) to improve both real and apparent age estimation, while other cases were dedicated to optimising the apparent age only. In both scenarios (Fig.~\ref{error:real-curve} and Fig.~\ref{error:app-curve}), the curves show lower error values for those ages with higher number of samples on the train set (Fig.~\ref{hist:train-set}).

\begin{figure}[htbp]
\centering
\subfigure[G$_{a|r}$ = 22.5 $|$ \textbf{26}, P$_{a|r}$ = 27.1 $|$ \textbf{26}]{\includegraphics[height=2.4cm]{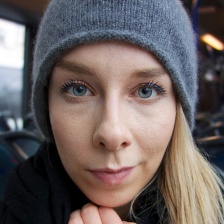}}\hspace{0.3cm}
\subfigure[G$_{a|r}$ = 42.5 $|$ \textbf{45}, P$_{a|r}$ = 43.3 $|$ \textbf{45}]{\includegraphics[height=2.4cm]{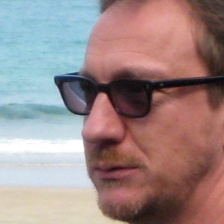}}\hspace{0.3cm}
\subfigure[G$_{a|r}$ = 29.7 $|$ \textbf{29}, P$_{a|r}$ = 30.1 $|$ \textbf{29}]{\includegraphics[height=2.4cm]{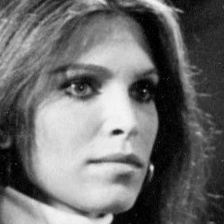}}
\subfigure[G$_{a|r}$ = 25.1 $|$ \textbf{24}, P$_{a|r}$ = 26.4 $|$ \textbf{24.1}]{\includegraphics[height=2.4cm]{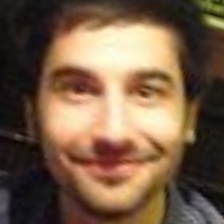}}\hspace{0.3cm}
\subfigure[G$_{a|r}$ = 76.4 $|$ \textbf{64}, P$_{a|r}$ = 58.2 $|$ \textbf{64.1}]{\includegraphics[height=2.4cm]{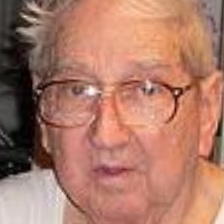}}\hspace{0.3cm}
\subfigure[G$_{a|r}$ = 7.6 $|$ \textbf{8}, P$_{a|r}$ = 4.0 $|$ \textbf{7.9}]{\includegraphics[height=2.4cm]{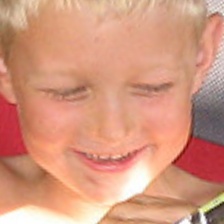}}
\subfigure[G$_{a|r}$ = 62.8 $|$ \textbf{56}, P$_{a|r}$ = 51.4 $|$ 
\textbf{56.4}]{\includegraphics[height=2.4cm]{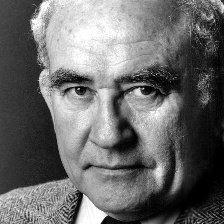}}\hspace{0.3cm}
\subfigure[G$_{a|r}$ = 87.4 $|$ \textbf{80}, P$_{a|r}$ = 72.0 $|$ \textbf{79.8}]{\includegraphics[height=2.4cm]{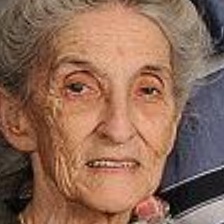}}\hspace{0.3cm}
\subfigure[G$_{a|r}$ = 26.3 $|$ \textbf{29}, P$_{a|r}$ = 28.3 $|$ \textbf{25.9}]{\includegraphics[height=2.4cm]{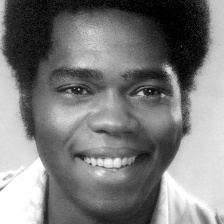}}
\caption{Qualitative results based on real age estimation error. G~=~ground truth, P~=~prediction, ($a|r$) = apparent or real age.}
\label{fig:results-1}
\end{figure}

\begin{figure}[htbp]
\centering
\subfigure[G$_{a|r}$ = \textbf{29.6} $|$ 33, P$_{a|r}$ = \textbf{29.6} $|$ 27.2]{\includegraphics[height=2.4cm]{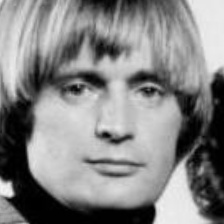}}\hspace{0.3cm}
\subfigure[G$_{a|r}$ = \textbf{1.2} $|$ 4, P$_{a|r}$ = \textbf{1.1} $|$ 6]{\includegraphics[height=2.4cm]{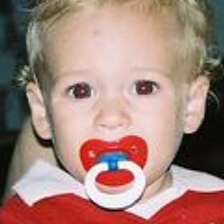}}\hspace{0.3cm}
\subfigure[G$_{a|r}$ = \textbf{30.1} $|$ 46, P$_{a|r}$ = \textbf{30.2} $|$ 30.8]{\includegraphics[height=2.4cm]{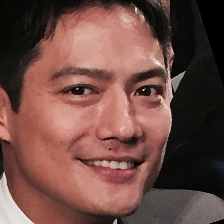}}
\subfigure[G$_{a|r}$ = \textbf{43.4} $|$ 46, P$_{a|r}$ = \textbf{43.3} $|$ 44.9]{\includegraphics[height=2.4cm]{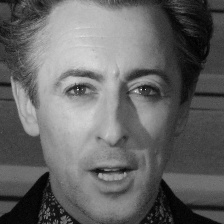}}\hspace{0.3cm}
\subfigure[G$_{a|r}$ = \textbf{28.5} $|$ 34, P$_{a|r}$ = \textbf{28.4} $|$ 27.4]{\includegraphics[height=2.4cm]{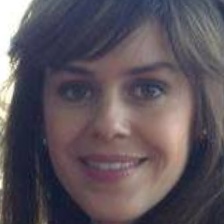}}\hspace{0.3cm}
\subfigure[G$_{a|r}$ = \textbf{14.5} $|$ 18, P$_{a|r}$ = \textbf{14.4} $|$ 13.6]{\includegraphics[height=2.4cm]{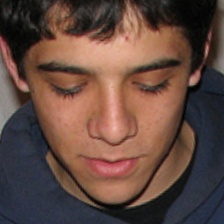}}
\subfigure[G$_{a|r}$ = \textbf{57.0} $|$ 67, P$_{a|r}$ = \textbf{56.7} $|$ 63.6]{\includegraphics[height=2.4cm]{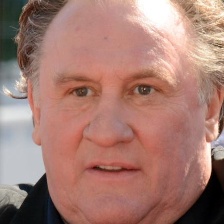}}\hspace{0.3cm}
\subfigure[G$_{a|r}$ = \textbf{18.7} $|$ 15, P$_{a|r}$ = \textbf{18.1} $|$ 16.2]{\includegraphics[height=2.4cm]{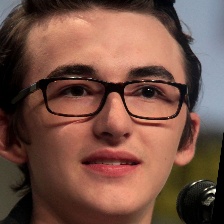}}\hspace{0.3cm}
\subfigure[G$_{a|r}$ = \textbf{11.2} $|$ 11, P$_{a|r}$ = \textbf{11.9} $|$ 12.4]{\includegraphics[height=2.4cm]{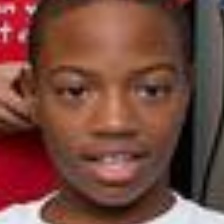}}
\caption{Qualitative results based on apparent age estimation error. G~=~ground truth, P~=~prediction, ($a|r$) = apparent or real age.}
\label{fig:results-2}
\end{figure}

\begin{figure}[htbp]
\centering
\subfigure[G$_{a|r}$ = 62.5 $|$ 60, P$_{a|r}$ = 30.2 $|$ 27.9]{\includegraphics[height=2.4cm]{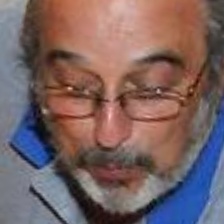}}\hspace{0.3cm}
\subfigure[G$_{a|r}$ = 6.3 $|$ 8, P$_{a|r}$ = 36.5 $|$ 35.7]{\includegraphics[height=2.4cm]{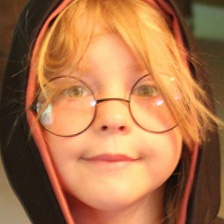}}\hspace{0.3cm}
\subfigure[G$_{a|r}$ = 83.9 $|$ 89, P$_{a|r}$ = 38.3 $|$ 39.3]{\includegraphics[height=2.4cm]{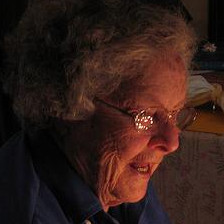}}
\caption{Unsatisfactory results obtained using the proposed model. G~=~ground truth, P~=~prediction, ($a|r$) = apparent or real age. }
\label{fig:results-3}
\end{figure}

Fig.~\ref{fig:results-1} shows qualitative results obtained using the proposed model, with images sorted based on real age estimation error. Additional qualitative results are shown in Fig.~\ref{fig:results-2}, in this case with images sorted based on apparent age estimation error. In Fig.~\ref{fig:results-3} we show few examples of unsatisfactory results for both cases, which may be caused due to partial occlusion, illumination condition, head-pose or even due small number of samples in the train set for those age ranges. 

%
%
%
%
\vspace{1.0cm}
%
%
%
%
%

\vspace{-1cm}

\subsection{Attribute-based analysis}\label{sec:attributes}


In Table~\ref{table:attributes} we show computed errors for each attribute and case study with respect to real and apparent age estimation. Results reported for Cases 1 and 2 do not use real age labels (\textit{i.e.}, the predicted real ages were based on apparent age labels). The percentile of each category on the train set is shown to illustrate the distribution of each attribute and the presence (or not) of unbalanced data. Note that all attributes have been analysed individually in order to make the analysis simple so that few and consistent observations could be taken. A higher level analysis could consider the intersection among different attributes, but probably would require a larger (and balanced) dataset.

\begin{table}[htbp]
\caption{Attribute analysis for real/apparent age estimation based on the error obtained on the test set for the three case studies (C1, C2 and C3). Results in bold represent lowest error rate for each type of prediction (real/app) and attribute.}
\label{table:attributes}
\vspace{-0.2cm}
\begin{center}
\begin{tabular}{|c|c|c|c|c|c|c|} \hline
\textbf{Att.} & \textbf{Pred.} & \textbf{\% Tr.} & \textbf{Category}  & \textbf{C1} & \textbf{C2} & \textbf{C3} \\ \hline
\multirow{4}{*}{Gender} & \multirow{2}{*}{Real} & 50.72 & Male & 8.29 & 6.63 & \textbf{6.55} \\ \cline{3-7}
                        & & 49.28 & Female & 10.05 & 8.28 & 8.11 \\ \cline{2-7}
                        & \multirow{2}{*}{App} & \multirow{2}{*}{''} & Male & 7.46 & 6.06 & 6.27 \\ \cline{4-7}
                        & & & Female & 7.59 & 5.99 & \textbf{5.99} \\ \hline \hline
\multirow{6}{*}{Race} & \multirow{3}{*}{Real} & 10.43 & Asian & 8.27 & 6.83 & \textbf{6.59} \\ \cline{3-7}
                        & & 86.6 & Caucasian & 9.25 & 7.51 & 7.40 \\ \cline{3-7}
                        & & 2.97 & Afroamerican & 9.65 & 8.12 & 7.73 \\ \cline{2-7}
                        & \multirow{3}{*}{App} & \multirow{3}{*}{''} & Asian & 7.12 & \textbf{5.24} & 5.36 \\ \cline{4-7}
                        & & & Caucasian & 7.58 & 6.10 & 6.21 \\ \cline{4-7} 
                        & & & Afroamerican & 6.93 & 5.34 & 5.30 \\ \hline \hline
\multirow{8}{*}{Happy} & \multirow{4}{*}{Real} & 17.53 & Happy & 9.28 & 7.85 & 7.58 \\ \cline{3-7}
                        & & 43.71 & Slightly & 9.67 & 7.86 & 7.63 \\ \cline{3-7}
                        & & 34.67 & Neutral & 8.86 & \textbf{6.98} & 6.99 \\ \cline{3-7}
                        & & 4.09 & Other & 8.98 & 7.28 & 7.34 \\ \cline{2-7}
                        & \multirow{4}{*}{App} & \multirow{4}{*}{''} & Happy & 7.35 & 6.08 & 6.11 \\ \cline{4-7}
                        & & & Slightly & 7.50 & 5.99 & 6.05 \\ \cline{4-7} 
                        & & & Neutral & 7.67 & \textbf{5.94} & 6.16 \\ \cline{4-7} 
                        & & & Other & 7.82 & 6.35 & 6.35 \\ \hline \hline
\multirow{8}{*}{Makeup} & \multirow{4}{*}{Real} & 19.72 & Makeup & 9.30 & 7.66 & 7.35 \\ \cline{3-7}
                        & & 72.33 & No makeup & 9.05 & 7.32 & 7.33 \\ \cline{3-7}
                        & & 0.98 & Not clear & 10.86 & 9.44 & 8.86 \\ \cline{3-7}
                        & & 6.98 & Very subtle & 9.96 & 7.77 & \textbf{7.20} \\ \cline{2-7}
                        & \multirow{4}{*}{App} & \multirow{4}{*}{''} & Makeup & 6.40 & 4.77 & \textbf{4.61} \\ \cline{4-7}
                        & & & No makeup & 8.10 & 6.66 & 6.92 \\ \cline{4-7} 
                        & & & Not clear & 8.66 & 6.38 & 5.76 \\ \cline{4-7} 
                        & & & Very subtle & 7.66 & 6.19 & 6.52 \\ \hline
\end{tabular}
\end{center}
\end{table}

From Table~\ref{table:attributes} we can make the following observations:
\begin{itemize}
\item \textbf{Gender}: real age is slightly better predicted for males in all cases, even though using a balanced train set. Regarding apparent age, obtained results are similar, with a slight error decrease for females.  \item \textbf{Race}: the train set for this attribute is strongly unbalanced. Taking real age estimation, in general, Afro-Americans obtained the highest real age estimation error, followed by Caucasians and Asians. Clearly, the small number of samples in the Afro-American category is the main reason for these results. Interestingly, Asians obtained the lowest real age error even when having less than 11\% of samples for training. Regarding apparent age, Afro-Americans obtained overall lowest error rates even with less than 3\% of samples on the train set. On the other hand, Caucasians obtained the highest error rates when having more than 86\% of samples for training. These results indicate that the perception of age of Asian and Afro-American people is more accurate if compared to Caucasians, which may reflect some physiological phenomena. For instance, it was related in~\cite{Vashi:2016} that ethnicity and skin colour have many characteristics that make its ageing process unique, and those of Asian, Hispanic, and African American descent have distinct facial structures.
\item \textbf{Happiness}: the presence of happiness (or ``other'' emotion category) demonstrated to negatively influence real age estimation. With respect to apparent age, no significant difference was observed. In general, neutral faces helped to improve overall results. These findings are aligned with~\cite{Voelkle:2012}, \textit{i.e.}, neutral faces are more easily estimated, whereas age of happy faces tends to be underestimated.
\item \textbf{Makeup}: even though having an unbalanced dataset for this attribute, no significant difference was observed for real age estimation (\textit{i.e.}, if ``not clear'' class is ignored, as it has less than 1\% of train data). In general, the absence of makeup (or very subtle makeup) helped to achieve lowest error rates for real age estimation. On the other hand, the presence of makeup helped to improve results for apparent age estimation.
\end{itemize}

\subsection{How your age is perceived based on the gender of the observer?}\label{sec:workers}

This section presents a proof of concept application to regress apparent age based on the input gender of a given observer. For this, we base on the gender information of labellers provided for a subset of the APPA-REAL dataset~\cite{clapes2018apparent}. Thus, the model can be trained considering gender observer information as an additional input variable to previously considered attributes. The main aim of this application is to automatically respond the question: how a male/female will perceive the age of the person in this image? Note that higher analysis considering the intersection among all attributes (observer \textit{vs.} observed people) could be considered as future work.

\begin{figure}[htbp]
      \centering
      \includegraphics[height=2.8cm]{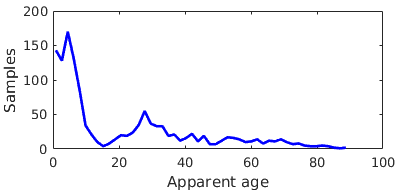}
      \caption{Apparent age distribution on the ``new train set'', \textit{i.e.}, when considering the gender of the observers.}
      \label{hist:app:worker}
   \end{figure}

For a subset of individuals on the train/validation sets, the gender of the observer is provided in addition to the respective apparent age label. For the test set, all samples contain annotations from different observers (of both genders, male/female). Then, for each sample on the train/validation/test set, we computed the average apparent age with respect to these two points of views, \textit{i.e.}, male/female observer. Thus, we have for each individual his/her average apparent age from two observer genders (which composes the ground truth of this ``new dataset''). It resulted in a small train set (compared to the original one) but large enough to report some initial results. In total, the new train set is composed of $1328$ samples and $504$ validation samples. Fig.~\ref{hist:app:worker} shows the distribution of this new train set.  

We modified the \textit{input\_2} layer of the proposed model (shown in Fig.~\ref{overview:proposed}) to include the gender of the observer, encoded as one hot vector and concatenated with the attributes of the person being observed. It resulted in a vector of length $15$ (instead of the original $13$). The model is trained as described in Sec.~\ref{sec:training}, using $lr_1=1e-4$.

\begin{figure}[htbp]
\centering
\subfigure[G$_{f|m}$=\{51.4~$|$~47.3\} P$_{f|m}$=\{51.0~$|$~50.54\}]{\includegraphics[height=2.8cm]{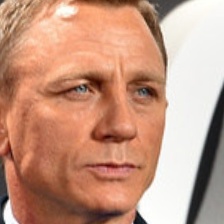}} \hspace{0.5cm}
\subfigure[G$_{f|m}$=\{29.1~$|$~27.4\} P$_{f|m}$=\{27.1~$|$~26.7\}]{\includegraphics[height=2.8cm]{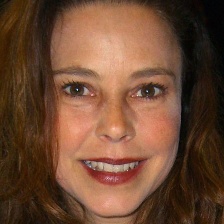}} \hspace{0.5cm}
\caption{Results obtained using the proposed model when taking into account the gender of the observer. G~=~ apparent age ground truth, P~=~apparent age prediction, ($f|m$) = female/male observer.}
\label{fig:results-worker}
\end{figure}

Obtained results indicate that the inclusion of the gender of the observer did not have a strong impact on the outcomes, as the computed error (MAE) for both points of views (males/females) are somehow similar, \textit{i.e.}, $9.758$ for female observers and $9.243$ for males. However, these preliminary results suggest that the male's perception was modelled with slightly better accuracy. Note that these results cannot be compared to those reported before since data and annotations have been modified. Some qualitative results obtained by the model are shown in Fig.~\ref{fig:results-worker}. Note that although differences in apparent ground truth for males and females do not highly differ, predictions of our model when input gender varies provides a closer apparent prediction consistent to the gender of the observer.


\section{Conclusion}\label{sec:conclusions}

In this work, we proposed an end-to-end CNN architecture for real age estimation that benefits from apparent age information. The network in a first stage uses face attributes in target image as input variables to learn their age perception biases and regress the apparent age. The second part of the network benefits from both input attributes of face and apparent age prediction to regress a final real age prediction, like performing unbias. We showed in the APPA-REAL dataset that proposed network, integrating both apparent and real age predictions, achieves better recognition for both tasks that when they are addressed separately. We also provided with a proof of concept application where the network was trained including the gender of the age guessers. This way, during testing we could retrieve the apparent age for an input image given the gender of an observer. 

Future work will include the extension of both amount of data and number of attributes for a deeper analysis of the bias involved in age perception. We plan to jointly recognise those attributes together with apparent and real age tasks for a fine-grain analysis on the problem, allowing the multi-task network to share weights from early training stages among all tasks.

\section*{Acknowledgements}
This work has been partially supported by ICREA, under the ICREA Academia programme, as well as by the Spanish projects TIN2015-66951-C2-2-R, TIN2016-74946-P (MINECO/FEDER,UE) and CERCA Programme/Generalitat de Catalunya, and by the Scientific and Technological Research Council of Turkey (TUBITAK) 1001 Project (116E097), and by the Estonian Centre of Excellence in IT (EXCITE) funded by the European Regional Development Fund. The authors also gratefully thank the support of NVIDIA Corporation with the donation of the GPUs used for this research.

\bibliographystyle{IEEEtran}


\end{document}